%% file: main.tex
\documentclass[conference,letterpaper]{IEEEtran}
\usepackage{booktabs}
\usepackage{graphicx}
\usepackage{todonotes}
\usepackage[export]{adjustbox}
\usepackage[caption=false,font=footnotesize]{subfig}
\usepackage{cite}

\usepackage{multirow} 

\usepackage[hidelinks]{hyperref}
\title{Cross-Lingual SynthDocs: A Large-Scale Synthetic Corpus for Any to Arabic OCR and Document Understanding}

\author{
\IEEEauthorblockN{
Haneen Al-Homoud, Asma Ibrahim, Murtadha Al-Jubran, Fahad Al-Otaibi,\\
Yazeed Al-Harbi, Daulet Toibazar, Kesen Wang, and Pedro J. Moreno}
\IEEEauthorblockA{
Humain \\ Riyadh, Saudi Arabia \\
\{hhomoud, aaibrahim, maljubran, ffalotaibi, yaharbi, dtoibazar, kwang, pmoreno\}@humain.ai}
}

\begin{document}
\maketitle
\input{sec/0_abstract}    
\input{sec/1_intro}
\input{sec/2_related_work}
\input{sec/3_methodology}

\input{sec/4_results_and_discussion}
\input{sec/5_conclusion}

\bibliographystyle{IEEEtran}
\bibliography{main}

\end{document}

%% file: sec/0_abstract.tex
\begin{abstract}

Cross-Lingual SynthDocs is a large-scale synthetic corpus designed to address the scarcity of Arabic resources for Optical Character Recognition (OCR) and Document Understanding (DU). The dataset comprises over 2.5 million of samples, including 1.5 million textual data, 270K fully annotated tables, and hundred thousands of real data based charts. Our pipeline leverages authentic scanned backgrounds, bilingual layouts, and diacritic aware fonts to capture the typographic and structural complexity of Arabic documents. In addition to text, the corpus includes variety of rendered styles for charts and tables. Finetuning Qwen-2.5-VL on SynthDocs yields consistent improvements in Word Error Rate (WER) and Character Error Rate (CER) in terms of OCR across multiple public Arabic benchmarks, Tree-Edit Distance Similarity (TEDS) and Chart Extraction Score (CharTeX) improved as well in other modalities. SynthDocs provides a scalable, visually realistic resource for advancing research in multilingual document analysis.
\end{abstract}

%% file: sec/1_intro.tex
\section{Introduction}
\label{sec:intro}
Arabic Document Understanding (DU) is extremely low-resourced compared to widely studied languages such as English, primarily due to the unavailability of high-quality annotated datasets and intrinsic linguistic and structural difficulties peculiar to Arabic script, including complicated morphology, optional diacritics, Right-To-Left (RTL) direction, and profuse typographic variations \cite{zaghouani2016building, el2021arabic}. In contrast, English benefits from abundant benchmark datasets, sophisticated annotation tools, and extensive research attention, which has enabled accelerated development in Optical Character Recognition (OCR) and DU tasks \cite{harley2015evaluation, mathew2021docvqa}. Furthermore, Arabic documents typically possess complex and varied layouts characterized by multi-column layouts structures, inset graphical elements, and non-sequential text flows, which further discourages machine-based text recognition and semantic extraction \cite{alyafeai2021arabert}. This absence of large, general Arabic corpora significantly holds up the development, benchmarking, and deployment of effective OCR and DU systems and thus brings progress in pivotal real-world applications such as automatic information retrieval, smart document analysis, and cross-lingual knowledge extraction \cite{obeid2020camel}.
 
Although previous efforts have attempted to tackle this problem, existing Arabic-focused datasets such as CAMeL Tools \cite{obeid2020camel}, Arabert\cite{alyafeai2021arabert}, and Khatt \cite{mahmoud2014khatt} are either tiny in size, strongly biased toward handwritten material compared to printed documents, or lack the extensive annotations necessary for sophisticated DU tasks. Additionally, even though there are cross-lingual synthetic corpora such as SynthDoc \cite{ding2024synthdoc} and DocSynth \cite{kil2021docsynth}, they have predominantly focused on European languages, in particular English, and lack the linguistic and structural realism required for high-quality Arabic DU tasks. These systems, while practical, skirt around key layout complexities, linguistic quirks, and morphological richness of Arabic, undermining their usefulness and applicability for Arabic OCR and DU scenarios.
 
To bridge this fundamental resource gap, we introduce Cross-Lingual SynthDocs, a novel large-scale synthetic corpus for enhancing Arabic OCR and DU tasks. By leveraging automatic text-generation methods, multilingual alignment methods, and realistic layout simulations, our system exploits the availability and richness of existing English and European language resources to create complete and high-quality Arabic annotations. This new cross-lingual approach effectively reduces the necessity for scarce human-annotated Arabic bibliographic information, enabling faster progress in Arabic OCR and DU technologies and narrowing the resource gap existing in Arabic document understanding research.

%% file: sec/2_related_work.tex
\section{Related Work}
\label{sec:related_work}

\subsection{Arabic OCR and Synthetic Document Generation}
Recent Arabic OCR research has achieved notable progress but still faces major challenges, largely due to limited access to large, high-quality annotated datasets. The SARD dataset \cite{ahmed2024sard}, for example, provides 843K synthetic Arabic document images – mainly from printed books, increasing the availability of training data and typographic variety. Yet progress remains limited; current models continue to struggle with realistic performance, especially on noisy inputs and complex, dynamic layouts. Approaches like Arabic-Nougat \cite{rashad2024arabic} and QARI-OCR \cite{wasfy2025qari} have used large multimodal language models and synthetic data to reduce the Character Error Rate (CER) to ~0.06 and improve layout sensitivity. However, their reliance on clean synthetic datasets still restricts effectiveness in real-world OCR scenarios, where irregular structures and noise dominate.

\subsection{Traditional OCR Approaches and Their Limitations}
Conventional Arabic OCR systems such as Tesseract \cite{smith2007overview} and classic Convolutional Neural Network Long Short-Term Memory (CNN–LSTM) frameworks have demonstrated effectiveness in recognizing both printed and handwritten scripts. Yet, they face persistent challenges: they are highly sensitive to noisy inputs, often mishandle the RTL reading order intrinsic to Arabic \cite{facchin2020whole}, and frequently misrepresent ligature formation, all of which are critical for accurate interpretation.
Classical post-processing methods, like dictionary checks and statistical language models, work on simple linear text \cite{aldahoul2023survey}. But they fail on documents with tables, sidebars, or overlapping text, because they cannot handle complex layouts. This is a big problem since nonlinear structures are very common in documents.

\subsection{Advances in Synthetic Data Generation for OCR}
Obtaining high-quality datasets is difficult, especially for low-resource languages like Arabic. Synthetic data models such as SynthDoc and DocSynth \cite{ding2024synthdoc, kil2021docsynth} work well for European languages and can generate flexible, realistic documents. But their performance drops sharply on Arabic because of its complex morphology, RTL writing, and frequent use of diacritics. These features make it hard to create realistic Arabic documents, and the output often misses important linguistic details. This shows the need for synthetic data methods designed specifically for Arabic and its unique structure.

\subsection{Post-OCR Correction and Layout Preservation}
Current OCR post-correction methods, such as weakly supervised error injection and test-time adaptation, improve accuracy on simple, linear text \cite{guan2024effective}. But they often fail on complex Arabic documents where text is spread across multiple lines and sections, making it hard to keep meaning and structure.
New spatial-domain clustering methods, like hybrid spatial-semantic chunking and density-based clustering, show promise by reasoning over local text features such as proximity, size, and overlap \cite{verma2025s2chunking,mcinnes2017hdbscan}. Still, they struggle with noisy input and unstable layouts, since traditional layout analysis cannot adapt well to real-world variation \cite{binmakhashen2019survey}. In short, even with recent progress, current solutions remain too rigid to handle dynamic, real-world document structures.


%% file: sec/3_methodology.tex
\section{Methodology}
\label{sec:methodology}
To address these gaps, our proposed Intelligent OCR Text Replacement Engine offers a solution for Arabic synthetic data generation and OCR post-processing. The system uses spatial grouping algorithms to cluster noisy bounding boxes into meaningful text regions and applies intelligent text allocation to place Arabic translations optimally within them. It also incorporates adaptive RTL rendering, including font-based glyph filtering, precise Unicode bidirectional markers, and irregular scaling. These steps ensure semantic consistency, visual coherence, and layout preservation even in noisy OCR settings. By focusing on issues often overlooked by existing systems. Such as noise robustness, RTL ordering, semantic consistency, and layout preservation, our method improves both the quality and the practical applicability of Arabic OCR synthetic data and editing.
In this section, we detail the techniques involved in our synthetic framework.
\label{sec:synthetic-dataset-text}

\subsection{Cropped Text Snippets}
We built an automated pipeline to produce cropped Arabic text samples with controlled diacritization. Using the \texttt{arbml/tashkeela} corpus as input, we applied two transformations: (1) selective removal of diacritics at light, medium, or heavy levels, and (2) random insertion of diacritics to increase variability. We also place Eastern Arabic and Western numerals at configurable frequencies to better reflect real usage.

The resulting text chunks were rendered into images with \LaTeX{} using authentic Arabic fonts. We varied fonts, sizes, and alignments across samples, and automatically cropped whitespace to ensure clean outputs.


\subsection{Synthetic Any to Arabic Dataset Creation}
To support robust evaluation of multilingual document understanding, we built an end-to-end synthesis pipeline that proceeds as follows.
\subsubsection{IDL Documents} A corpus from the Industry Documents Library (IDL), a public archive of over three million documents spanning a diverse range of business, legal, and technical reports \cite{pixparse_idl_wds, biten2022ocr}. Each IDL document comprises a PDF alongside a JSON file of line-level OCR annotations.


\subsubsection{Paragraph Grouping via Adjacency Graph}
Although the IDL JSON annotation captures every line of text with its bounding box and transcription, these raw lines do not correspond to semantically coherent paragraphs. To reconstruct natural reading units, we build an adjacency graph in which each node represents one OCR line. Two nodes are connected if their baselines are separated by no more than the page's median line spacing (computed dynamically per page) and their horizontal extents overlap by at least 30\% of the upper line's width. We then extract connected components from the graph: each component corresponds to one contiguous paragraph region.

\subsubsection{Translation and Layout Preservation}
Once paragraphs are identified, we reassemble the lines in various languages into coherent text blocks and submit each block to Gemini 2.0 Flash for machine translation \cite{geminiteam2025geminifamilyhighlycapable}. To maintain fidelity, we enforce two key constraints on the model’s output: abbreviations (e.g.,\ “U.S.”, “Inc.”) and symbol tokens (e.g.,\ “\$”, “\%”) must remain unchanged, and where the original lines were discontiguous (e.g.,\ list items or headings), the translation API should return separate translated lines rather than a merged paragraph. After receiving the Arabic (or mixed Any–Arabic) translation, we redistribute the translated sentences back into their original bounding boxes, adjusting text direction to RTL.

\subsubsection{Rendering Synthetic Pages}
With translated text and original layout metadata in hand, we render fully synthetic pages using Python's Pillow library. For each page, we randomly sample typefaces from a curated set of Arabic and Latin fonts, along with complementary color palettes, to simulate real-world document styles. We respect each box's text direction, which is RTL for Arabic language, and apply proper alignment. We then composite the translation atop the original page background, faithfully reproducing table gridlines, embedded charts, and decorative elements.  





\subsection{Table Data}
\label{sec:table-data}

We have created a large dataset of HTML tables to help with Arabic DU tasks like detecting tables, recognizing structures, and OCR. Our dataset consists of two main sources of tables: arxiv and synthetic tables.

\subsubsection*{ArXiv tables} - We collected \LaTeX{} source files through the arXiv API and extracted table environments enclosed between \verb|\begin{table}| and \verb|\end{table}|. Since the original language of arXiv documents is English, we applied an Large Language Model (LLM)-based content modification step: the structural layout of the tables was preserved from the original \LaTeX{} files, while the textual content was translated and adapted into Arabic. To achieve a standardized format, the standalone \LaTeX{} tables were converted into HTML tables, which provide greater flexibility for both structural alignment and content moderation. As content in arXiv papers generally appears with a white background and standard fonts, we re-rendered the final tables as images directly from the HTML representations.

\subsubsection*{Synthetic Tables} 
We constructed a synthetic collection of Arabic tables to bridge the existing gap in publicly accessible datasets. This collection consists of two distinct table categories:

\paragraph*{1- Consistent Style Tables – Realistic Layout Simulation} These tables are produced through automated processes designed to faithfully replicate the formatting structures characteristic of genuine Arabic documentation. Each table is designed to ensure consistency, legibility, and realism in representing document structures.

The tables are well organized, with regular rows and columns, and may include merged cells to represent grouped or hierarchical information. Header and footer rows are highlighted with background shading, while captions can appear at the top or bottom. A uniform style is maintained within each table, using consistent font sizes, text colors, and background shading, all rendered with authentic Arabic fonts. Text direction is set to RTL in line with Arabic writing, with variations in vertical alignment added for realism.

Consistency is also preserved at the content level: Arabic text may appear with or without diacritics, accompanied by either Eastern or Western numerals, while English tables use only Western numerals. Together, these design choices create tables that are both structured and realistic, closely reflecting real world document formatting.

\paragraph*{2- Random Style Tables - Augmentation for Robustness} The random style tables provide substantial visual variability that serves as an essential augmentation that greatly improves the model generalization on the dataset.

Some key attributes of this category are the high level of variation applied at the cell level. Font family, size, and color can differ from one cell to another, and background colors also vary across the table. In addition, cell content are randomly inserted into individual cells and rows. Horizontal and vertical alignments are randomized per cell, further increasing variability. Tables may also include unpredictable elements such as headers, footers, captions, and merged cells, similar to those found in real-world documents. Within a single cell, the content can combine Arabic text (with or without diacritics), text in other languages, and numeric values written in both Eastern and Western digits.

By combining these two types, the dataset captures clean, standardized tables and more complex, unstructured ones. This makes it useful for training layout aware models to better handle the diversity of real world Arabic document styles.

\subsection{Charts Data} 
The path we followed to generate the charts dataset began with reading real chart data points and then using an LLM to generate additional required values, including more complex cases such as three dimensional charts. Once the data points were prepared, an automatic generation pipeline was applied to produce thousands of charts in a wide variety of styles. This pipeline randomized visual elements such as themes, fonts, background, and text rotation, while ensuring that each chart was paired with structured annotations containing its title, type, and tabular data points in a CSV like format. This approach provided both realism and diversity.

%% file: sec/4_results_and_discussion.tex

\section{Results and Discussion}
\label{sec:results_and_discussion}
\subsection{Dataset}

In the following subsections, we describe our synthetic corpus for OCR and structure parsing. More samples of rendered text crops, pages and annotated tables and charts are available at  
\url{https://huggingface.co/datasets/Humain-DocU/SynthDocs}.

\subsubsection{Textual Data}

We generated a large scale Arabic corpus in two forms: page level renderings and cropped text snippets. 

We generated 1 million cropped PNG images of Arabic text, each paired with a matching plain text file. The images vary in font, size, and sentence length, with different levels of diacritization.
The text was rendered with a wide range of authentic Arabic fonts, including 
Amiri, Amiri Quran, FreeSerif, and many Kacst fonts 
(KacstFarsi, KacstDigital, KacstPen, KacstScreen, 
KacstOne, KacstArt, KacstTitle, KacstTitleL, KacstQurn, mry\_KacstQurn, KacstNaskh, KacstLetter, KacstOffice, KacstDecorative, KacstBook, KacstPoster).

This ensures coverage of traditional, decorative, and modern digital styles.

In addition to cropped text snippets, we built an automated pipeline to produce a full page dataset. We converted a substantial subset of the original \texttt{pixparse/idl‑wds} dataset into a large scale Arabic document corpus by replacing each page that is in a different language with its Arabic translation while preserving its rich structural and annotation metadata for OCR processing \cite{pixparse_idl_wds}.  The \texttt{pixparse/idl‑wds} collection comprising PDF files, TIFF renderings, and extensive JSON annotations provided an ideal foundation for our pipeline (Section~\ref{sec:synthetic-dataset-text}).  Applying that methodology, we processed 140,277 documents (570,636 pages).

To simulate the typographic diversity of real world Arabic documents and improve model robustness, we randomized text appearance across 15 carefully selected font families and nine distinct text colors.  Our font palette spans:
\begin{itemize}
  \item Traditional calligraphic styles: ArefRuqaa, ReemKufi, Saudi
  \item Contemporary sans‑serif designs: Cairo, Fund
  \item Monospaced faces: AzarMehr, MonospacedSans, AzarMehrMonospacedSerif, KawkabMono, DejaVuSansMono
  \item Specialized display fonts: TheYearoftheCamel, MusmekSaudi, NaseebSaudi, WatadSaudi
  \item System defaults: Courier, SimSunExtB
\end{itemize}
This systematic variation produced a visually authentic synthetic dataset that better approximates the heterogeneity of real world Arabic document layouts.

\subsubsection{Tabular Data}
\label{sec:Tabular Data}

\begin{figure*}[t]
  \centering
  \subfloat[\label{fig:a-table}]{
    \includegraphics[frame,width=0.29\textwidth]{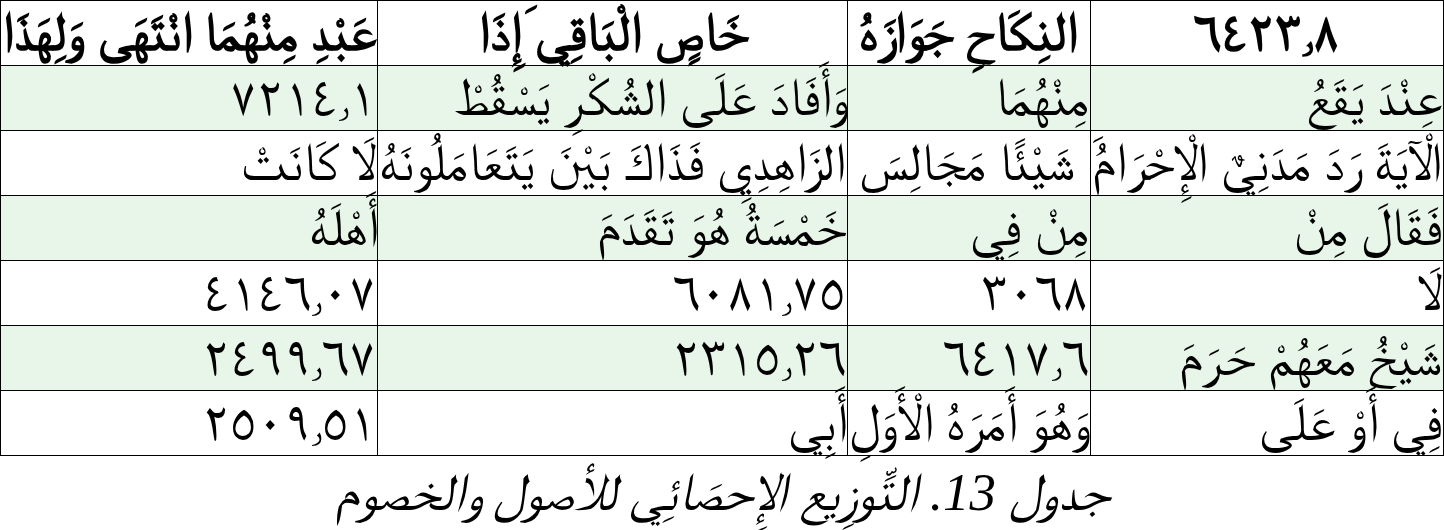}
  }
  \hfill
  \subfloat[\label{fig:b-table}]{
    \includegraphics[frame,width=0.29\textwidth]{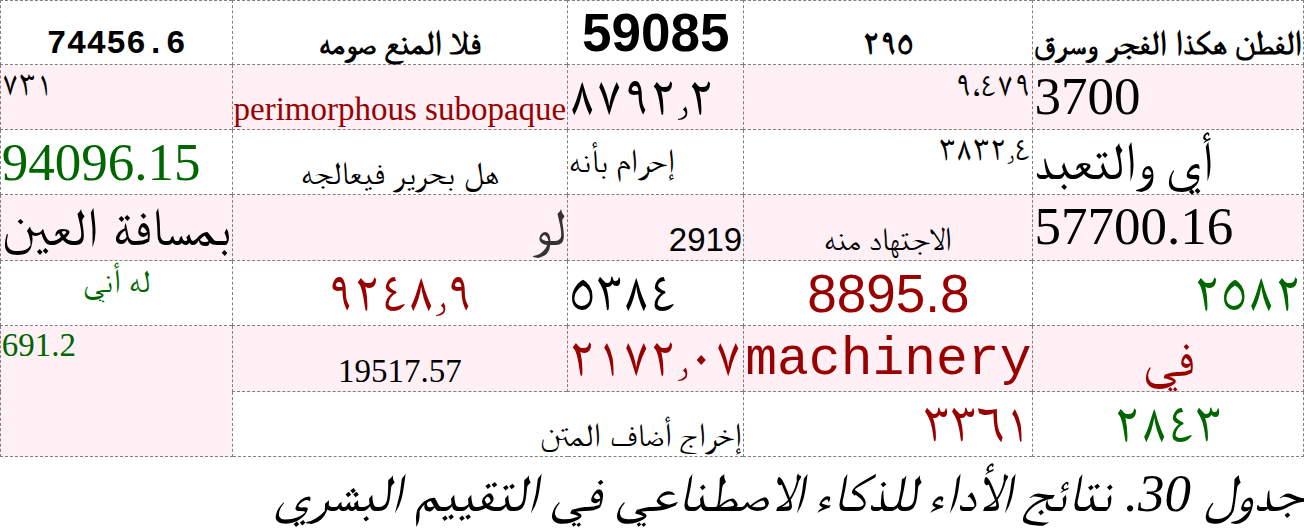}
  }
  \hfill
  \subfloat[\label{fig:c-table}]{
    \includegraphics[frame,width=0.29\textwidth]{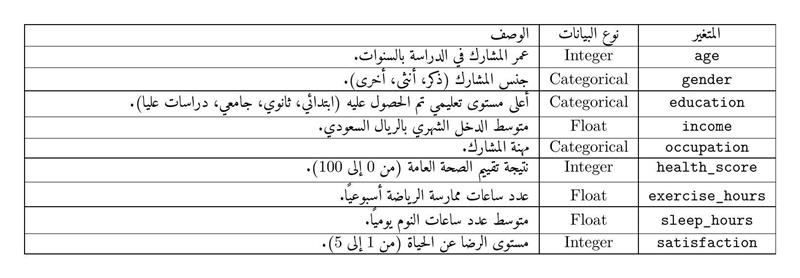}
  }
    \caption{Examples of synthetic tables: (a) a consistent style table with uniform formatting, (b) a random style table with varied fonts and content, and (c) a table generated from the ArXiv synthetic subset.}
  \label{fig:tables-examples}
\end{figure*}

Our synthetic table dataset comprises 270,000 tables,  designed to support Arabic DU capabilities such as table detection, table structure recognition, and structured HTML-parsing. We integrated these tables with synthetic data, thereby combining real world table layouts with artificially generated content. The synthetic table data set contains two generation strategies:

\begin{itemize}
\item The 200,000 tables follow a consistent style, reflecting real world formatting and layout found in Arabic or bilingual documents such as reports and statements. These tables are neat and professional looking like a formal document. Fig.~\ref{fig:a-table} shows examples of a consistent style table.

\item The 50,000 tables were generated using the random style formatting, which added inconsistency to layout, noisy formatting, and some tables even contained mixed languages. This portion was interpreted as a data augmentation strategy to allow models to practice with Out-OfDistribution (OOD) variations, and could improve generalization. Fig.~\ref{fig:b-table} illustrates the rendering of a random style table.

\item The 20,000 tables were generated from the ArXiv API explained in Section~\ref{sec:table-data}, an example of a generated table is shown in Fig.~\ref{fig:c-table}.
\end{itemize}

We keep the ground truth table content in HTML form so that we can have a consistent representation of structure as well as content. To bring tables data in sync with downstream vision language model (VLM) training, we focus on a table to HTML parsing objective. We normalize every HTML sample prior training as following: 1) The removal of redundant wrappers and tags. 2) Keeping only the table structure and actually visible content. This reduction makes sure that the model will learn to extract the minimal layout and semantic structure without being overfit to noise or boilerplate HTML structures.

With our highly efficient generation pipeline, it is effortless to scale the tabular data to millions without dependency on any LLMs.

\subsubsection{Chart Data}
\begin{figure*}[t]
  \centering
  \subfloat[\label{fig:0196}]{
    \includegraphics[frame,width=0.24\textwidth]{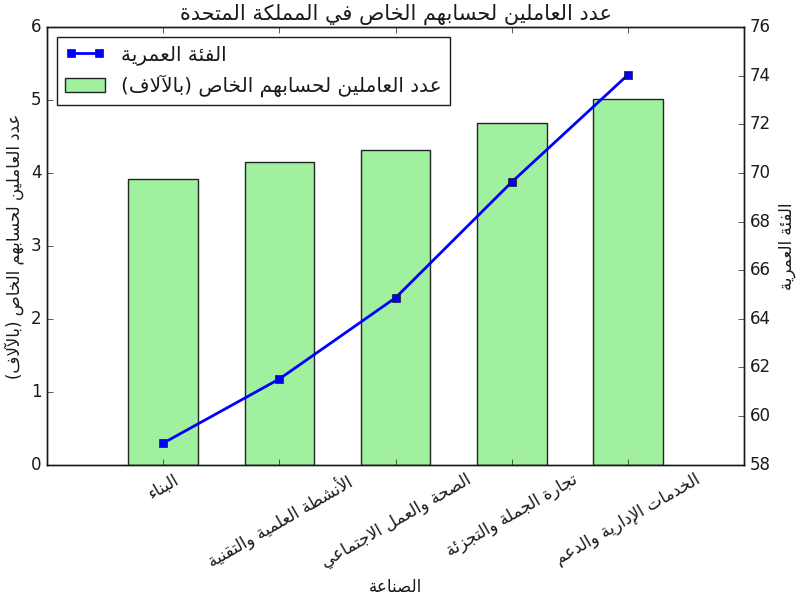}
  }
  \subfloat[\label{fig:0557}]{
    \includegraphics[frame,width=0.24\textwidth]{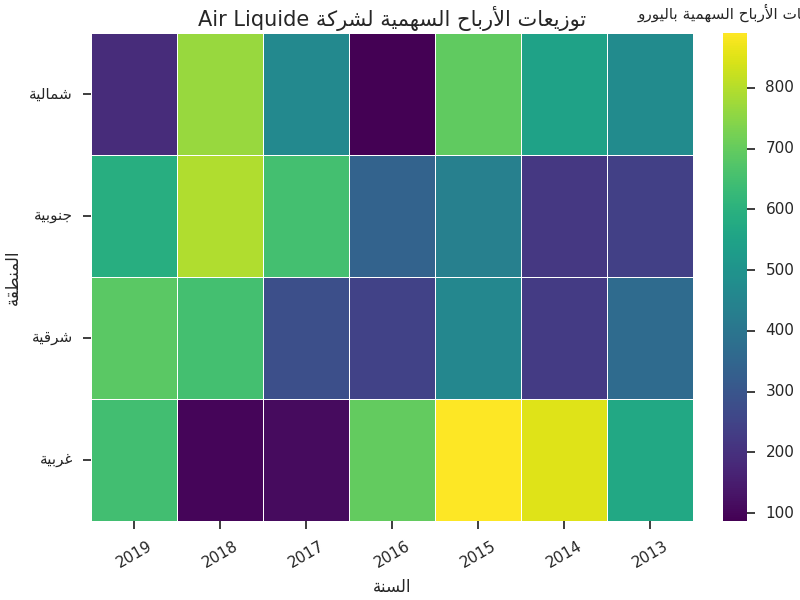}
  }
  \subfloat[\label{fig:area}]{
    \includegraphics[frame,width=0.24\textwidth]{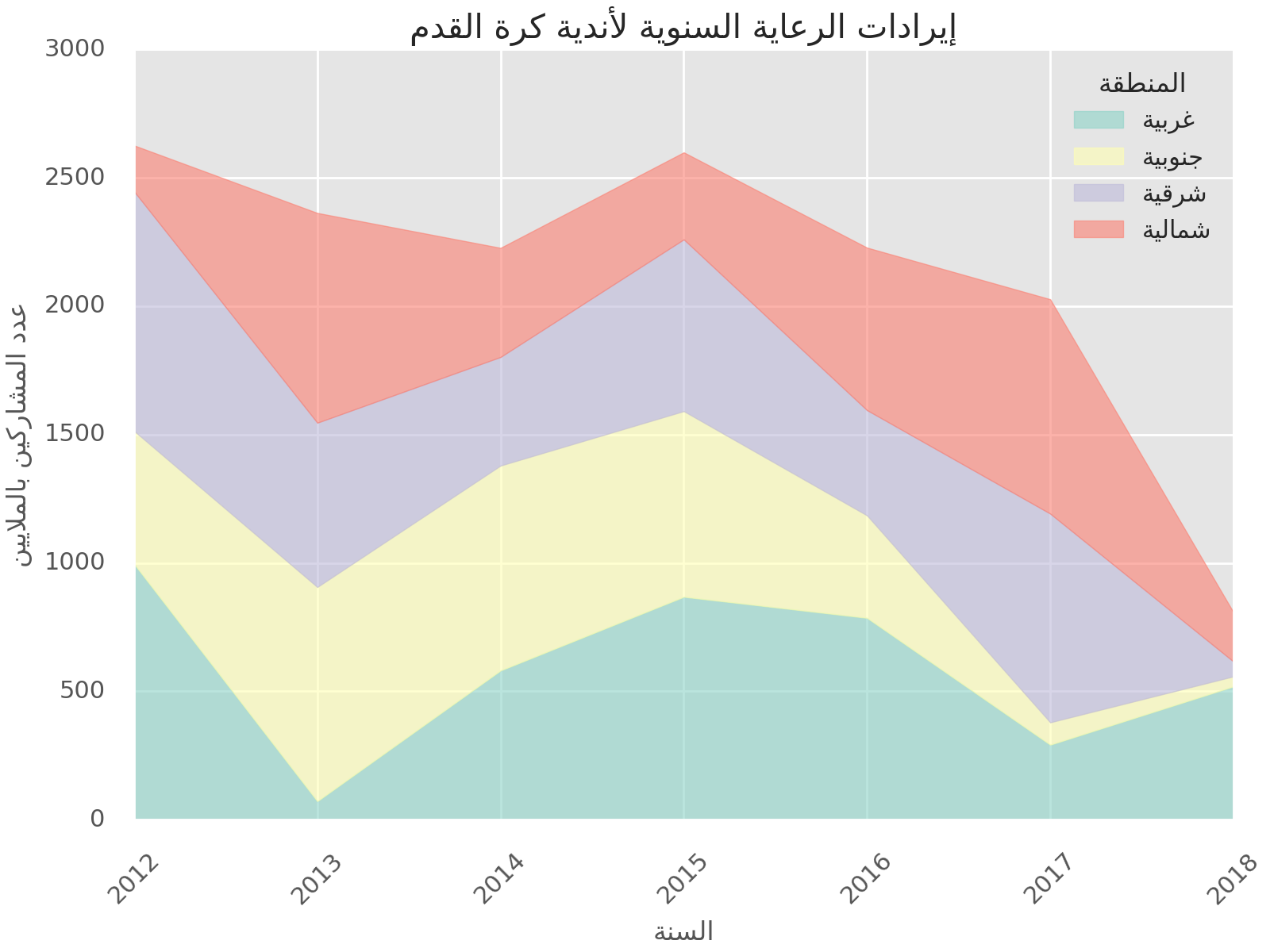}
  }
  \subfloat[\label{fig:ring}]{
    \includegraphics[frame,width=0.24\textwidth]{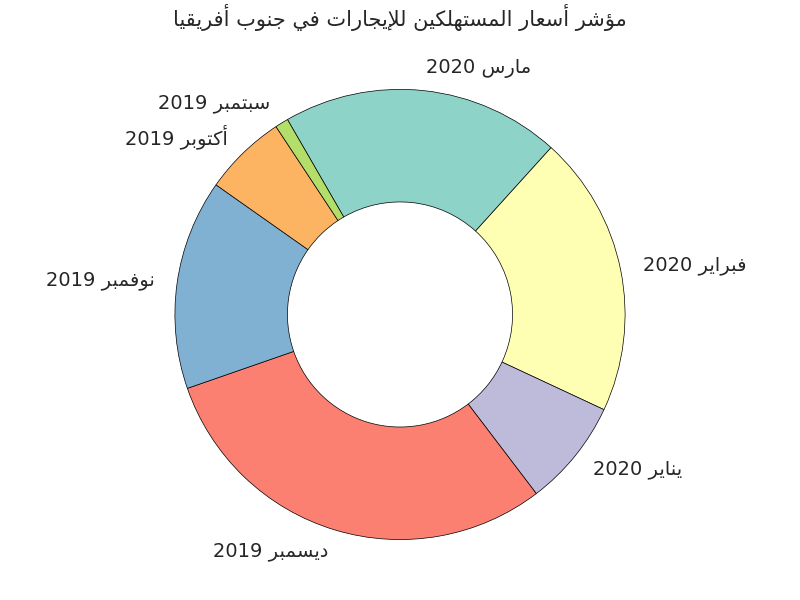}
  }
  \caption{Examples of synthetically generated charts: (a) Dual-axis chart, (b) Heatmap, (c) Area chart, and (d) Doughnut chart.}
  \label{fig:charts}
\end{figure*}

The final dataset consists of 600,000 charts covering 15 standard chart types, including pie, bar, grouped bar, stacked bar, line, area, dot, histogram, scatter, box, violin, heatmap, dual-axis, doughnut, and bubble. Each chart is accompanied by structured annotations that capture its metadata title, type and its data points. Fig.~\ref{fig:charts} shows examples of generated charts, including dual-axis, heatmap, area, and doughnut styles.

\subsection{OCR Model Experiments}
\subsubsection{Text}
We use the OCR benchmarks from KITAB-Bench \cite{heakl2025kitab} to evaluate the performance of our finetuned Qwen-2.5-VL-3B, a Large Vision Language Model (LVLM), on the synthetic corpus generated by our system.

Results in Table~\ref{tab:ocr_results_transposed} demonstrate that the finetuned Qwen-2.5-VL-3B yields consistently substantial gains across diverse Arabic OCR benchmarks. Numbers in bold in the table highlight where the finetuned model surpasses its base counterpart. Performance is measured using both Word Error Rate (WER), which captures the proportion of words that are incorrectly predicted relative to the reference text, and Character Error Rate (CER), which analogously measures errors at the character level.

\begin{table*}[t]
  \centering
  \footnotesize
  \caption{OCR performance of LVLMs on Arabic benchmarks (WER/CER as percentages). Lower is better.}
  \label{tab:ocr_results_transposed}
  \begin{tabular}{@{}lccc@{}}
    \toprule
    \textbf{Dataset} &
    \textbf{\shortstack{Gemini-2.0-Flash \\(WER - CER)}} &
    \textbf{\shortstack{Qwen-2.5-3B \\(WER - CER)}} &
    \textbf{\shortstack{Qwen-2.5-3B Finetuned \\(WER - CER)}} \\
    \midrule
    ArabicOCR         & 2.0\%  -- 0.0\%   & 8.0\%  -- 2.0\%   & \textbf{2.0\%  -- 0.0\%} \\
    PatsOCR           & \textbf{2.0\%  -- 1.0\%}   & 37.0\% -- 27.0\%  & 10.0\% -- 6.0\% \\
    SynthesizedAr     & 22.0\% -- 8.0\%   & 52.0\% -- 37.0\%  & \textbf{20.0\% -- 6.0\%} \\
    ISI-PPT           & 21.0\% -- 7.0\%   & 97.0\% -- 86.0\%  & \textbf{17.0\% -- 7.0\%} \\
    Local Arabic Set        & 13.7\% -- 7.7\%   & 20.9\% -- 10.7\%  & \textbf{12.8\% -- 5.1\%} \\
    Local English Set        & 5.1\%  -- \textbf{1.6\%}   & 4.9\%  -- 2.6\%   & \textbf{5.0\%}  -- 2.0\% \\
    \bottomrule
  \end{tabular}
\end{table*}

From clean to structured Arabic (PatsOCR \& ArabicOCR). Beginning with cleaner, more regular text, finetuning on our set drives sharp error reductions: we reach parity with Gemini-2.0-Flash on the cleaner multi-line set and essentially catch up on the more varied single-line set. These gains suggest improved ligature handling and robust glyph normalization across fonts.

Stepping up to style/noise stress (SynthesizedAr \& ISI-PPT). As we move to heavier diacritization, missing characters, colored/noisy backgrounds, and cross-lingual content, performance transitions from unreliable to competitive. Errors more than halve overall, surpassing Gemini on the style-heavy synthetic set and closing most of the gap on slide-derived text indicating stronger shape invariance and resilience to degraded typography.

Generalization. Beyond the public benchmarks, Arabic on our in house data improves materially and exceeds Gemini, while English maintains WER parity with a small CER gain, evidencing cross-lingual transfer without forgetting. Taken together, finetuning with our synthetic corpus yields large relative WER reductions (up to \(\sim 80\%\)), approaches Gemini on clean text, and surpasses it on challenging, style-intensive Arabic scenarios.

\subsubsection{Charts}
To evaluate the effectiveness of our chart dataset, we finetuned Qwen2.5-VL-7B on the synthetic charts and assessed performance on the KITAB-Bench chart parsing benchmark using the Chart Extraction Score (CharTeX) \cite{heakl2025kitab}. Table~\ref{tab:parsing_results} reports the comparison against baseline LVLMs.

Among the baselines, Gemini-2.0-Flash achieved the strongest performance with a score of 53.49. The base Qwen2.5-VL-7B model, without finetuning, underperformed at 18.65, highlighting its limited ability to parse charts. After finetuning on our synthetic charts dataset, Qwen2.5-VL-7B improved substantially to 48.97, a gain of over 30 points.

This result demonstrates that targeted finetuning with our large scale synthetic charts dataset significantly boosts chart understanding and parsing capabilities, narrowing the gap with state of the art LVLMs. It also highlights the value of the dataset as an effective resource for improving charts-specific reasoning and data extraction in Arabic and multilingual document settings.

\begin{table}[t]
  \centering
  \small
  \caption{Parsing results on the KITAB bench dataset across different LVLMs. Results are reported on charts using CharTEx and on tables using TEDS.}
  \label{tab:parsing_results}
  \begin{tabular}{@{}lllc@{}}
    \toprule
    \textbf{Task} & \textbf{Metric} & \textbf{Model} & \textbf{Score (\%)} \\
    \midrule
    \multirow{4}{*}{Charts} & \multirow{4}{*}{CharTEx} 
      & GPT-4o-mini                & 41.15 \\
    & & Gemini-2.0-Flash           & 53.49 \\
    & & Qwen2.5-VL-7B              & 18.65 \\
    & & \textbf{Qwen2.5-VL-7B Finetuned} & \textbf{48.97} \\
    \midrule
    \multirow{4}{*}{Tables} & \multirow{4}{*}{TEDS} 
      & GPT-4o-mini                & 69.32 \\
    & & Gemini-2.0-Flash           & 83.86 \\
    & & Base Qwen2.5-VL-32B        & 81.65 \\
    & & \textbf{Qwen2.5-VL-32B Finetuned} & \textbf{83.68} \\
    \bottomrule
  \end{tabular}
\end{table}

\subsubsection{Tables}
Table parsing results were evaluated using the Tree Edit Distance Based Similarity (TEDS) metric~\cite{DBLP:journals/corr/abs-1911-10683}. 
As shown in Table~\ref{tab:parsing_results}, Gemini-2.0-Flash achieved the strongest performance with a TEDS score of 83.86. 
The base Qwen-2.5-VL-32B model obtained 81.65, but after finetuning on our diverse tabular dataset comprising both synthetic tables and real tables extracted from reports it improved to 83.68, a relative gain of about 2 points.  

Although the absolute gain may appear modest, it is significant in the context of structural understanding. KITAB-Bench \cite{heakl2025kitab} tables often present multi-row headers, merged cells, and inconsistent numeral styles, creating substantial difficulties for parsing. The finetuned model demonstrates improved robustness in recovering structural and semantic information from such complex tables.  

These results highlight the effectiveness of augmenting training with synthetic tables, showing that our dataset contributes meaningfully to the advancement of table understanding in Arabic document analysis.


%% file: sec/5_conclusion.tex
\section{Conclusion}
\label{sec:conclusion}
In this work, we introduced Cross-Lingual SynthDocs, a large-scale synthetic dataset designed to address the shortage of Arabic resources for OCR and DU. The dataset combines over 2.5 million samples, including rendered pages, annotated crops, annotated tables and charts, all built with realistic  fonts, layouts, diacritic and different styles variations. By leveraging automatic generation and cross lingual alignment, we created data that captures the complexity of real Arabic documents.

Our experiments show that finetuning Qwen on SynthDocs leads to consistent improvements across multiple Arabic OCR and structure parsing benchmarks. In particular, the models achieved significant reductions in both CER and WER and improvements in TEDS and CharTEx matrices, while also narrowing the gap with SOTA systems like Gemini. These results confirm that synthetic, large scale, and visually realistic resources can play a key role in advancing Arabic OCR and document understanding.

Looking ahead, we plan to extend SynthDocs with further modalities (e.g., form synthesis, info-graphics), integrate real world distortions for robustness, and explore domain adaptation and broader LVLM benchmarking.